\documentclass{llncs}		          %[proof]
\usepackage{proof}			          %\infer
\usepackage{fixltx2e} 		        %\textsubscript
\usepackage{mathtools}			      %\vcentcolon \tag
\usepackage[utf8x]{inputenc}      %$∑$
\usepackage{xcolor}			          %\color
\usepackage[labelfont=bf,labelsep=period,singlelinecheck=false]{caption}	%specify bold, .-sep-d, left-aligned captions
\usepackage{cite}                 %nice cites
\usepackage{enumerate}            %(i)
\usepackage{float}                %tables
\usepackage{mathpazo}             %use hott font
\usepackage{hyperref}	            %clickable x-ref-s (load this pack last; [hidelinks])
\renewcommand{\t}{\thinspace}
\renewcommand{\d}{\vcentcolon=}
\newcommand{\n}{\noindent}
\newcommand{\N}{\mathbb{N}}
\newcommand{\Z}{\mathcal{Z}}
\newcommand{\M}{\mathcal{M}}
\newcommand{\R}{\mathcal{R}}

\begin{document}

\title{Modeling selectional restrictions in a relational type system}
\author{Erkki Luuk\thanks{I thank Zhaohui Luo for discussions and comments on earlier versions of this material. The work has been supported by IUT20-56 and European Regional Development Fund through CEES.}}
\institute{Institute of Computer Science, University of Tartu, Estonia \\
\email{erkkil@gmail.com}}

\maketitle

\begin{abstract}
Selectional restrictions are semantic constraints on forming certain complex types in natural language. The paper gives an overview of modeling selectional restrictions in a relational type system with morphological and syntactic types. We discuss some foundations of the system and ways of formalizing selectional restrictions. 
\end{abstract}

%artikkel alaku

\section{Background}\label{sec:bg}

A type theory, whether simple or complex, is essential for a logical specification of natural language (NL). In its complex form (i.e. as dependent and/or polymorphic type theory), it is the most expressive logical system (contrasted with the nonlogical ones such as set and category theory). As \cite{asher_selectional_2014,luo_semantics_2010,luo_formal_2014,ranta_type-theoretical_1994} have shown, in a logical approach (i.e. in one with simpler alternatives such as zeroth, first, second and higher order logic), complex type theories outshine simpler ones in accounting for phenomena like anaphora, selectional restrictions, etc. Also, as the notion of type is inherently semantic:
\begin{enumerate}[(i)]
\item \label{i} type $\d$ category of semantic value,
\end{enumerate}
it is by definition suited for analyzing universal phenomena in NL, as NL semantics in largely universal (as witnessed by the possibility of translation from any human language to another).

Type-theoretical modeling of NL has a long background, modeling selectional restrictions less so. The latter topic, nearly overlooked in Montagovian\cite{montague_proper_2002} and categorial\cite{lambek_mathematics_1958} traditions, has been investigated only recently \cite{asher_selectional_2014,luo_semantics_2010}. The logical essence of the linguistic phenomenon of selectional restrictions is fixing types for a function's (or relation's\footnote{I will henceforth refer to them as ``relations" (as functions are relations).}) arguments. However, put in this way, the notion is not even informally precise. There is a difference between (1) arguments conforming to selectional restrictions and (2) relations imposing selectional restrictions to their arguments. Clearly, (1) and (2) are not mutually exclusive, but if we consider (1) in isolation, it seems natural to admit that an argument can conform to different selectional restrictions. For example, the first argument of \emph{read} might be a physical (Phy) and sentient entity, while the second might be a physical and informational entity. On the other hand, if we see selectional restrictions as being imposed by the relations, different selectional restrictions for one argument position become less viable. One reason is that multiple restrictions for an argument position cannot be \emph{directly} computed by a function. The second one is that multiple restrictions for an argument position suggest type polymorphism, which (for some reason or another) may seem undesirable. For example, \cite{asher_selectional_2014}-style approach would not work with multiple types per relation's argument position.

There are several ways out of these two near-plights. As different propositions may hold per position, the first problem can be avoided by indirect computations returning truth values. Alternatively, the computations might return sets of restrictions instead of restrictions. The second problem can be diverted with a type combining the properties of different types (e.g. with a $∑$- or subset type). This approach (with dot-types instead of $∑$-types) has been pioneered in \cite{luo_semantics_2010}. Upon a closer inspection, however, the idea of relations imposing different restrictions per positions does not seem attractive. Besides the abovementioned (and other) complications, it seems unnecessary. For example, since most sentient (like most informational) entities are also physical, the hypothesis of restriction Phy being imposed by \emph{read} can be substituted with that of a statistical correlation. Indeed, counterexamples, e.g. \emph{reading ones mind/thoughts} or \emph{a program reading a data} are available. On the other hand, a program can be considered a physical object in computer memory and/or the counterexamples can be dismissed altogether as derived metaphoric uses. However, metaphors are pervasive in language, so it makes sense merely to say that a use is \emph{more} metaphoric than another, thus effacing a clear distinction between the original and derived.

\section{An overview of $\Z$}\label{sec:oz}

In this paper, I develop some ideas about selectional restrictions in a relational model of NL. In this framework, the basic unit of NL is a relation of a finite (usually very small, $\leq14$) arity. Call the type system we are considering $\Z$. There are many universes in $\Z$, some of which are listed in table \ref{mu}:

\begin{table}[H]                          %\usepackage{float} + [H] fixes table position
\caption{Main universes}\label{mu}        %\label must follow \caption
\medskip
$\M \d$ the universe of morphosyntactic types (relations)\\
$\R \d$ the universe of selectional restrictions\\
Phy, Inf, Ani, Sen, Cou, Mas... : $\R$\\
Phy $\d$ physical entity\\
Inf $\d$ informational entity\\
Ani $\d$ animate entity\\
Sen $\d$ sentient entity\\
Cou $\d$ countable entity\\
Mas $\d$ mass entity\\
\end{table}

\n $\Z$ is a relational type system for modeling NL syntax, morphology and compositional semantics. The system has no proper terms, its lowest-order elements are types. The main universe in $\Z$ is that of morphosyntactic relations $\M$. We can distinguish the subuniverse of $n$th-order types $\M_n$ in $\M$, as well as subuniverses for any particular value of $n$. Note that $\M_n$ is not a lower-order universe wrt. $\M$, i.e. we do not have $\M_n:\M$, nor do we have $\M_n:\M_{n+m}$ for any $m, n$. Morphemes, words, phrases and sentences are $\M_n$-types. Starting from the bottom, we have the following rules.\\

\n Morphemes as types:

\begin{equation}\label{map}
\infer[\mbox{MAP}]{a:\M_n}{a ~\mbox{morpheme}}
\color{white}{\tag*{MAP}}
\end{equation}

\n \ref{map} has a(n) (dis)advantage. The advantage is that it captures the natural alignment between definition \eqref{i} and the definition of morphemes as smallest meaningful NL units. The disadvantage is that invoking propositions-as-types on \ref{map} suggests an unnatural alignment of linguistic categories (morphemes are not propositions in NL). However, morphemes are abstractions that can be interpreted as propositions that morphemes are inhabited by morpheme instances (or uses), so an abstract propositions-as-types interpretation is not barred for morphemes.

Some morphemes are 0th-order $\M$-types ($\M_0$-types). An $\M_0$-type is one that occurs only in an argument position. Not all morphemes are $\M_0$-types. For examples, stems have order order 0, while plural markers, as relations over stems, have order 1. $\M_n$-types are ``natural" interpretations of particular morphemes, words, phrases and sentences as linguistic expressions\footnote{Whenever a value is needed for $n$, it usually turns out to be very small, $\leq14$.}. We claim that they are interpreted as certain relations (e.g. a word is interpreted as a relation involving the morphemes that it is composed of, a phrase is interpreted as a relation involving the words that it is composed of, etc.). This corresponds to the usual principle of compositionality, as it is known since Frege. Higher-order types (or universes) are linguistic categories that these interpretations inhabit, e.g. V, N, NP, ACC, etc. (cf. Tab. \ref{tab:eu}).\\

\n Complex $\M_n$-type formation:

\begin{equation}\label{nf}
\infer[\mbox{$n$-Form ($n\geq1$)}]{a(\bar{e}):\M_n}{\mbox{c}(a,\bar{e}) \mapsto a(\bar{e})\qquad a:\M_n}
\color{white}{\tag*{$n$-Form ($n\geq1$)}}
\end{equation}

\n $\bar{e} \d$ a possibly empty finite sequence of types $e_1, ..., e_n$\\
$x(y) \d$ a relation $x$ over $y$\\
c$(a,\bar{e}) \d$ an admissible concatenation of $a, e_1, ..., e_n$ in a NL\\
$\mapsto \t \d$ a parsing function\\

\n \ref{nf} conveys that if an admissible concatenation c$(a,\bar{e})$ is parsed as $a(\bar{e})$, $a(\bar{e})$ has type $\M_n$. For every $a(\bar{e})$, if the highest-order type among $e_1, ..., e_n$ has order $n$, $a$ and $a(\bar{e})$ have order $n$+1. We conflate c$(a,\bar{e})$ with a relation formula in $\M_n$. The idea is that \emph{red book}, \emph{livre rouge}, etc., are relation formulas in idiosyncratic notations (viz. English, French, etc.). As linguistic expressions are relation formulas over $\M_n$, they are naturally parsed as relations.

How to decide whether a particular relation $a(\bar{e})$ holds? Usually, one has a basic intuition about what modifies what (modification is a subcase of relation). The main sources for the intuition are morpheme or word classes and semantic contribution tests. For example, \emph{-s} modifies (i.e. is a relation over) \emph{work} in \emph{works} rather than vice versa, as (1) affixes modify stems not vice versa, (2) person/tense and plural markers modify flexibles rather than vice versa, and (3) \emph{-s} contributes to the meaning of \emph{work} in \emph{works} rather than vice versa. By a similar argument, \emph{heavy} modifies \emph{rain} in \emph{heavy rain} rather than vice versa, \emph{sleeps} modifies \emph{john} in \emph{john sleeps} rather than vice versa, etc. In each case, there's a clear asymmetry between functions of the components, as conveyed by (1)-(2) the functions of word and morpheme classes and (3) semantic contribution tests.\\

\n Elementary universe formation:

\begin{equation}\label{euf}
\infer[\mbox{EU-Form}]{T:\M}{T ~\mbox{in Tab. \ref{tab:eu}}}
%\color{white}{\tag*{EU-Form}}
\end{equation}

\n Forming elementary universes requires a comprehensive list of morphosyntactic categories. Some elementary universes are listed in the table below:

\begin{table}[H]
\caption{Elementary universes}\label{tab:eu}
\medskip
A $\d$ adjective\\
ACC $\d$ accusative\\
ADL $\d$ adverbial\\
ADP $\d$ adposition\\
X $\d$ core argument (from XP)
\end{table}
\n Complex universe formation and elimination:

\[
\infer[\mbox{CU-Form}]{A(B, ...) : \M}{a(b_1, ...) : \M\qquad a:A:\M\qquad b_1:B:\M}
\]
\[
\infer[\mbox{CU-Elim}]{a_{r_1, ...}(b_1, ...) : \M}{a_{r_1, ...}:A:\M\qquad b_1:B:\M\qquad A(B, ...) : \M\qquad b_1:r_1:\R}
\]
\\
\n $a_{r_1, ...}(b_1, ...)$ should be read ``$a$ imposing selectional restriction $r_1$ on its argument $b_1$ (and possibly more restrictions on other arguments)". As all terms are types in $\Z$, $(b_1, ...)$ is an $n$-indexed sequence of arbitrary types (with $b_1$ the first argument and $B$ a universe it belongs to). $A$ is a universe of relations, $B$ a universe of its arguments. Thus it would not make sense to consider $A$ a relation over $B$ in $A(B, ...)$\iffalse
\footnote{Thanks to Zhaohui Luo for this observation.}\fi. The latter is a notation for a more complex type, which can be formalized in different ways (e.g. as a $∑$- or $∏$-type). The following examples may be useful for a better understanding of the rules:

\[
\infer{\mbox{A(X) type}}{\mbox{red(car) type\qquad red : A\qquad car : X}}
\]
\[
\infer{\mbox{red(car) type}}{\mbox{red\textsubscript{Phy} : A\qquad car : X\qquad A(X) type\qquad car : Phy}}
\]
\smallskip

\section{Formalizing selectional restrictions}\label{sec:sr}

\n We can compute selectional restrictions directly with a partial function

\begin{equation} \label{sr}
\mbox{s} : \M_n → \N → \R,
\end{equation}

\n and reason about them with predicates

\begin{equation} \label{sr0}
\mbox{p} : \R → \mbox{Prop},
\end{equation}

\n An important theorem about selectional restrictions is
%\thickmuskip=0.5\thickmuskip
\begin{equation} \label{srt}
∏(x:\M_n)∏(y:\N).\t 1 \leq y \leq \mbox{ar}(y) ↔ \mbox{p}(\mbox{s}(x)(y)),
\end{equation}
%\thickmuskip=0mu

\n where ar : $\M_n → \N$ is the arity function. The theorem states that selectional restrictions are imposed on all suitable argument positions of all suitable types. The proof of the theorem is by cases (it has to be --- to ensure that all the suitable types have been set up correctly). Along with other similar theorems and lemmas, this has been formalized for a tiny (but interesting) fragment of $\M_n$\footnote{\url{http://ut.ee/~el/a/sr.v}}. Somewhat interestingly, all the main proofs there are identical. As alluded above, the main utility of the lemmas is to safeguard the formalization of morphosyntactic types and their selectional restrictions. In addition, they help to check answers to two interesting experimental questions, both dependent on the definition of `selectional restrictions': 1. Do all morphosyntactic types with arity $\geq 1$ project selectional restrictions? 2. Do morphosyntactic types that project selectional restrictions project them for all their argument positions?

\section{Conclusion}

We have discussed the role of selectional restrictions in a relational type system of morphosyntactic types. Our exposition differs in several aspects from those of \cite{asher_selectional_2014,luo_semantics_2010}. The main difference is that \cite{asher_selectional_2014,luo_semantics_2010} consider selectional restrictions as a purely semantic phenomenon, while the present model focuses on syntax (and morphology --- although I have not discussed the relationship between morphology and selectional restrictions, which is yet unclear). Selectional restrictions form an interface between syntax and semantics, guiding the formation of syntactic types from semantic consideration\cite{luuk_syntax-semantics_2015}. This investigation, itself a work in progress, is a part of a larger work in progress on type system $\Z$.

\bibliography{mybib}

\begin{thebibliography}{Mon02}

\bibitem[Ash14]{asher_selectional_2014}
Nicholas Asher.
\newblock Selectional restrictions, types and categories.
\newblock {\em Journal of Applied Logic}, 12(1):75--87, 2014.

\bibitem[Lam58]{lambek_mathematics_1958}
Joachim Lambek.
\newblock The mathematics of sentence structure.
\newblock {\em The American Mathematical Monthly}, 65(3):154--170, 1958.

\bibitem[Luo10]{luo_semantics_2010}
Zhaohui Luo.
\newblock Semantics and {Linguistic} {Theory} 20.
\newblock volume~20, pages 38--56, Vancouver, 2010.

\bibitem[Luo14]{luo_formal_2014}
Zhaohui Luo.
\newblock Formal semantics in modern type theories: is it model-theoretic,
  proof-theoretic, or both?
\newblock In Nicholas Asher and Sergei Soloviev, editors, {\em Logical
  {Aspects} of {Computational} {Linguistics} 2014 ({LACL} 2014)}, number 8535
  in {LNCS}, pages 177--188. Springer, Berlin, Heidelberg, 2014.

\bibitem[Luu15]{luuk_syntax-semantics_2015}
Erkki Luuk.
\newblock Syntax-semantics interface.
\newblock In {\em International {Encyclopedia} of {Social} and {Behavioral}
  {Sciences}}, pages 900--905. Elsevier, Oxford, 2015.

\bibitem[Mon02]{montague_proper_2002}
Richard Montague.
\newblock The proper treatment of quantification in ordinary {English}.
\newblock In Paul Portner and Barbara~H. Partee, editors, {\em Formal
  {Semantics}: {The} {Essential} {Readings}}, pages 17--34. Blackwell, Oxford,
  2002.

\bibitem[Ran94]{ranta_type-theoretical_1994}
Aarne Ranta.
\newblock {\em Type-theoretical grammar}.
\newblock Clarendon Press, Oxford; New York, 1994.

\end{thebibliography}

%\begin{thebibliography}{[MT1]}
%\end{thebibliography}

\end{document}